\PassOptionsToPackage{table,dvipsnames}{xcolor}
\documentclass[]{fairmeta}
\usepackage[utf8]{inputenc}
\usepackage{amsmath,amssymb,amsfonts,amsthm,mathtools}
\usepackage{enumitem}
\usepackage{listings}
\usepackage{inconsolata}
\usepackage{capt-of}
\usepackage{wrapfig}
\usepackage{pifont}
\usepackage{fontawesome5}
\usepackage{array}

\definecolor{cmarkgreen}{HTML}{006400} 
\definecolor{xmarkred}{HTML}{8B0000}   

\newcommand{\greencheck}{\textcolor{cmarkgreen}{\ding{51}}}
\newcommand{\redcross}{\textcolor{xmarkred}{\ding{55}}}
\newcolumntype{L}[1]{>{\raggedright\arraybackslash}p{#1}}
\newcommand{\mypar}[1]{\noindent \textbf{#1}}

\newcommand{\method}{SkillAligner}

\setlabdisplayname{OmniAI Group of ZJU ACES Lab}
\setuniversityname{}

\fancypagestyle{firstheader}{%
  \fancyhf{}%
  \fancyhead[R]{%
    \IfFileExists{others/ZJU.png}{\includegraphics[height=0.72cm]{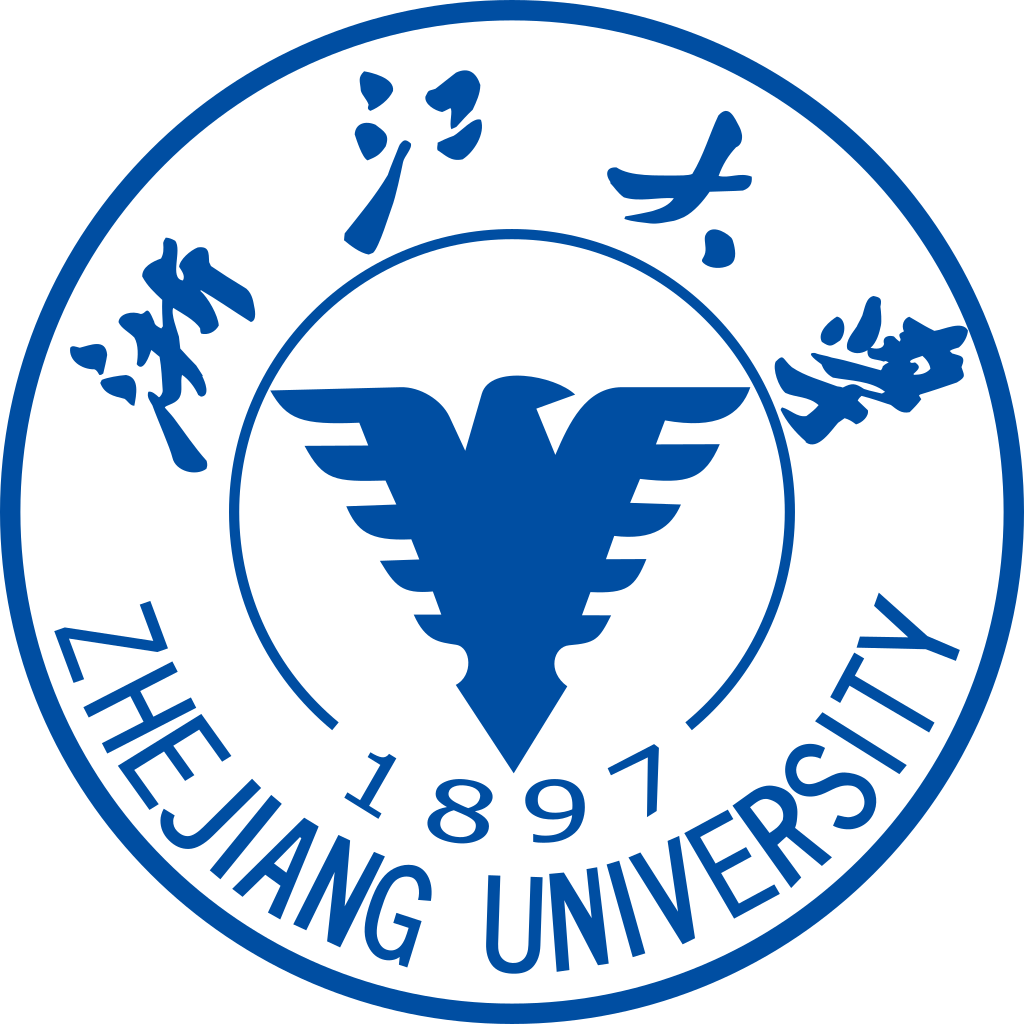}}{}%
  }%
}

\title{SkillAligner: Treating Retrieved Skills as Adaptable Drafts at Execution Time}

\author[*]{Qinfeng Li}
\author[*]{Dalin He}
\author[*]{Yuntai Bao}
\author{Ying Yang}
\author{Ruoxi Chen}
\author{Xinyan Yu}
\author{Lizhou Liang}
\author{Ge Su}
\author{Wenqi Zhang}
\author[\ddagger]{Xuhong Zhang}
\affiliation{Zhejiang University}
\contribution[*]{Equal contribution}
\contribution[\ddagger]{Corresponding author}
\metadata[\faEnvelope\ Email]{\email{zhangxuhong@zju.edu.cn}}
\date{August 2026}

\hypersetup{
  pdftitle={SkillAligner: Treating Retrieved Skills as Adaptable Drafts at Execution Time},
  pdfauthor={Qinfeng Li, DaLin He, Yuntai Bao, YingYang, Ruoxi Chen, Xinyan Yu, Lizhou Liang, Ge Su, Wenqi Zhang, Xuhong Zhang},
  pdfsubject={Training-free execution-time adaptation of retrieved procedural skills for language agents},
  pdfkeywords={language agents, procedural skills, skill adaptation, execution-time adaptation, skill-execution misfit}
}

\abstract{%
General-purpose skills promise reusable procedural knowledge for language agents, yet semantic relevance does not guarantee execution utility: a retrieved skill may encode assumptions that conflict with the current task, execution environment, or other retrieved skills. We formalize this problem as the \emph{skill--execution misfit}. To address it, we propose \textbf{\method{}}, a training-free execution-time skill adaptation framework that treats retrieved skills as adaptable drafts rather than fixed instructions. Before execution, \method{} performs a one-time joint adaptation that specializes useful skill fragments to task requirements, aligns their procedural assumptions with the available execution interface, and composes the resulting guidance by resolving dependencies, conflicts, and redundancy across skills. The adapted content is consolidated into a compact execution guide and reused throughout the subsequent trajectory. Extensive experiments across diverse agent benchmarks and model backbones show that \method{} substantially improves task performance over existing skill-use baselines, reduces skill-induced regressions at the instance level, and lowers total inference cost.
}

\begin{document}

\maketitle
\thispagestyle{firstheader}

\section{Introduction}
\label{sec:introduction}

Skill-augmented agents have emerged as a promising paradigm that organizes reusable procedural knowledge into structured skills~\citep{xia2026skillrl,wang2025sage,shi2026skill1,lin2026museautoskill}.
Following this paradigm, recent work has studied how to generate, store, retrieve, and evolve such reusable skills, thereby improving the availability of procedural knowledge for downstream agent execution~\citep{su2026skillretrieval,cho2026skillret,zhou2026skillgenbench}.

Existing work has primarily focused on obtaining high-quality skills~\citep{lin2026museautoskill,su2026skillretrieval}.
However, as these methods continue to develop, the bottleneck shifts from obtaining high-quality skills to ensuring their \emph{execution fitness}.
Specifically, retrieving a high-quality skill is not enough: it may still fail to fit the concrete execution and even degrade performance~\citep{skillsbench2026}.
To verify this phenomenon, we conduct a comparison by running each task both with and without relevant skills (as detailed in Section~\ref{Relevance--Utility_Gap} and Table~\ref{tab:relevance_utility_gap}).
We observe clear skill-induced regressions, where some tasks are solved without skills but fail after adding relevant skills.
We refer to this phenomenon as \emph{skill--execution misfit}: skills may fail to improve, or even harm, execution utility.
This raises a central question: \textbf{\textit{why do retrieved skills sometimes hurt execution, and how can we ensure the skills actually help agent execution?}}

\newpage
\begin{wrapfigure}[21]{r}{0.45\textwidth}
    \vspace{-0.75\baselineskip}
    \centering
    \includegraphics[width=\linewidth]{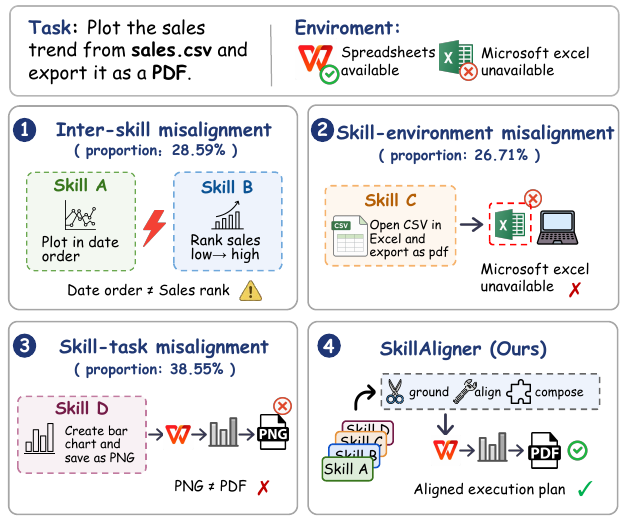}
    \caption{\textbf{Overview of the skill--execution misfit and \method{}.}
    Semantically relevant skills may still be misaligned with the task, environment, or other skills; \method{} addresses these mismatches through task grounding, execution-environment alignment, and skill composition.}
    \label{skill-use_misalignment}
    \vspace{-0.5\baselineskip}
\end{wrapfigure}

To identify the sources of \emph{skill--execution misfit}, we inspect execution trajectories from skill-induced regression cases (see Section~\ref{Relevance--Utility_Gap} and Table~\ref{tab:regression_sources}) and find that the failure modes can be categorized into three forms of misalignment. Specifically, as shown in Figure~\ref{skill-use_misalignment}, \emph{inter-skill misalignment} occurs when a skill cannot be coherently combined with other skills, for example, when one skill requires preserving chronological order while another requires sorting the same data by value. \emph{Skill--environment misalignment} occurs when a skill requires actions, resources, or states that are misaligned in the current execution environment. \emph{Skill--task misalignment} occurs when the skill's specific goal or constraints do not precisely match the requirements of the current task.


Unfortunately, existing methods remain insufficient to resolve \emph{skill--execution misfit}. \textbf{\textit{Their shared limitation is that skills are typically fully specified before the concrete execution instance is known.}} Specifically, \textbf{most methods} improve the quality of individual skills~\citep{shi2026skill1,su2026skillretrieval}, but do not account for conflicts, redundancy, or dependencies that emerge only when multiple skills are used together. \textbf{Structured composition methods} model such inter-skill relations through dependency-aware structures, such as graphs or skill bundles, thereby reducing \emph{inter-skill misalignment}~\citep{liu2026graphskills,xia2026grasp}. However, they primarily organize skills as fixed units and do not revise the procedural assumptions within each skill against the tools, actions, and resources available in the current execution, leaving \emph{skill--environment misalignment} unresolved. \textbf{Execution-grounded refinement methods} revise persistent skills using past execution traces~\citep{liu2026skillrevise,gautam2026skillaxe,gao2026skillaudit,lin2026museautoskill,yang2026skillopt}. While this improves compatibility with previously observed executions, the refined skill is still reused across future queries and therefore cannot directly specialize its procedures to instance-specific goals and constraints, leaving \emph{skill--task misalignment} unresolved.

Together, these limitations reveal a fundamental constraint of the prevailing paradigm: \textbf{\textit{skills are typically finalized before a concrete execution instance is known.}}
However, the user query, execution environment, and other retrieved skills may vary from one execution to another, making it difficult to determine in advance how a skill should be applied in every case.
As a result, the fixed skill may not match the current task, execution environment, or other skills, leading to three misalignments described above.
\textbf{\textit{Accordingly, our key insight is to treat retrieved skills as adaptable drafts rather than fixed artifacts and adapt them at execution time into guidance tailored to the current execution.}}

Based on this insight, we propose \textbf{\method{}}, an execution-time skill adaptation framework that converts retrieved skills into guidance tailored to the current execution. Specifically, for each query, \method{} first extracts a task contract and decomposes the retrieved skills into fragments, retaining task-consistent content while compressing or removing irrelevant fragments. It then checks the retained procedures against the available tools, action formats, resources, and permissions, repairing supported mismatches and converting uncertain procedures into conditional checks or fallbacks. Finally, it models dependencies, conflicts, and overlap among the retained fragments and composes them into a compact guide with four fields: \emph{Primary}, \emph{Checks}, \emph{Avoid}, and \emph{Fallback}. The resulting guide is used in place of the skills throughout the subsequent trajectory.

The \textbf{contributions} of our work are summarized as follows:
\begin{enumerate}[nosep, leftmargin=*]
\item We identify and formalize the \emph{skill--execution misfit} in skill-augmented agents: retrieved skills can harm execution. To diagnose this gap, we introduce \emph{skill-induced regression}, in which adding relevant skills turns a successful execution into a failure, and attribute these regressions to three sources of execution misalignment: inter-skill, skill--environment, and skill--task misalignment.
\item To the best of our knowledge, we present the \textit{first} framework for adapting retrieved general skills at execution time. \method{} filters task-irrelevant content, adapts skills to the available execution, and resolves conflicts across skills to produce a coherent execution guide.
\item Extensive experiments show that \method{} substantially improves task performance over existing skill-use baselines, reduces skill-induced regressions at the instance level, and lowers total inference cost.
\end{enumerate}


\begin{table}[t]
\centering
\setlength{\tabcolsep}{3.8pt}
\renewcommand{\arraystretch}{1.08}
\resizebox{.7\linewidth}{!}{%
\begin{tabular}{@{}lcccccc@{}}
\toprule
\multirow{2}{*}{\textbf{Benchmark}}
& \multicolumn{2}{c}{\textbf{Qwen3-8B}}
& \multicolumn{2}{c}{\textbf{Qwen3-32B}}
& \multicolumn{2}{c}{\textbf{Gemini 2.5 Pro}} \\
\cmidrule(lr){2-3}
\cmidrule(lr){4-5}
\cmidrule(lr){6-7}
& \redcross\,\(\rightarrow\)\,\greencheck~(\(\uparrow\))
& \greencheck\,\(\rightarrow\)\,\redcross~(\(\downarrow\))
& \redcross\,\(\rightarrow\)\,\greencheck~(\(\uparrow\))
& \greencheck\,\(\rightarrow\)\,\redcross~(\(\downarrow\))
& \redcross\,\(\rightarrow\)\,\greencheck~(\(\uparrow\))
& \greencheck\,\(\rightarrow\)\,\redcross~(\(\downarrow\)) \\
\midrule
ALFWorld
& 26.87 & \textbf{17.91}
& 20.90 & \textbf{14.93}
& 16.42 & \textbf{10.45} \\
WebShop
& 19.00 & \textbf{9.40}
& 14.20 & \textbf{6.60}
& 10.20 & \textbf{5.80} \\
SearchQA
& 14.52 & \textbf{10.29}
& 16.49 & \textbf{11.31}
& 14.71 & \textbf{11.48} \\
\midrule
Overall & 14.59 & \textbf{10.30} & 16.48 & \textbf{11.27} & 14.67 & \textbf{11.42} \\
\bottomrule
\end{tabular}
}
\caption{\textbf{Skill--execution misfit.}
We compare no-skill execution with direct top-$5$ raw skill injection on the same task instances.
Transfer (\redcross\,\(\rightarrow\)\,\greencheck) denotes instances that change from failure to success after adding skills, whereas
Regressions (\greencheck\,\(\rightarrow\)\,\redcross) denotes instances that change from success to failure.
The Overall row pools task instances.}
\label{tab:relevance_utility_gap}
\end{table}

\begin{table}[t]
\centering
\small
\setlength{\tabcolsep}{6pt}
\renewcommand{\arraystretch}{1.08}
\resizebox{.6\linewidth}{!}{%
\begin{tabular}{@{}lcccc@{}}
\toprule
Benchmark
& \textbf{Inter-skill}
& \textbf{Environment}
& \textbf{Task}
& Other \\
\midrule
ALFWorld & 34.84 & 23.81 & 30.36 & 10.99 \\
WebShop  & 28.60 & 24.64 & 39.30 & 7.46 \\
SearchQA & 22.34 & 31.67 & 46.00 & 0.00 \\
\midrule
Overall  & 28.59 & 26.71 & 38.55 & 6.15 \\
\bottomrule
\end{tabular}
}
\caption{\textbf{Sources of skill-induced regressions.}
We report the distribution of the primary mismatch among the regression cases included in source diagnosis.
The Overall row reports the macro-average across benchmarks.}
\label{tab:regression_sources}
\end{table}

\section{Preliminaries and Motivating Analysis}
\label{sec:prelim}

\subsection{Skill-Augmented Execution}

Skill-augmented agents solve tasks by conditioning on external reusable skills.
A skill encodes procedural knowledge, such as action steps, reasoning patterns, constraints, or domain-specific heuristics.
Given a task query, the system retrieves relevant skills from a skill library and provides them as additional guidance for planning and execution~\citep{xia2026skillrl,su2026skillretrieval}.
However, even semantically relevant skills are not necessarily execution-ready.
Their procedures and assumptions may conflict with the current task, execution environment, or other retrieved skills.
We refer to this incompatibility as \emph{skill--execution misfit}~\citep{skillsbench2026}.

\subsection{Diagnosing the Skill--Execution Misfit}
\label{Relevance--Utility_Gap}

\paragraph{Instance-level regression.}
We examine whether skill-induced regressions persist across benchmarks and backbone capabilities. For each backbone, we compare the same task instances with and without direct raw-skill injection under otherwise identical settings (detailed setup is provided in Appendix~\ref{app:relevance_utility_setup}). We define \emph{transfer} as an instance that fails without skills but succeeds after skill injection, and \emph{regression} as an instance that succeeds without skills but fails after skill injection. \textbf{\textit{Skill-induced regression is pervasive across benchmarks and backbone capabilities.}} As shown in Table~\ref{tab:relevance_utility_gap}, raw skills yield more transfers than regressions in all nine benchmark--backbone settings, but still induce regressions in every setting, ranging from 5.80\% to 17.91\%. \textbf{\textit{Notably, even stronger backbone models do not substantially alleviate this issue.}} Specifically, at the pooled level, Gemini 2.5 Pro (11.42\%) does not exhibit greater robustness to skill-induced regression than Qwen3-8B or Qwen3-32B, with regression rates of 10.30\% and 11.27\%, respectively.

\paragraph{Sources of regression.}
We further analyze why relevant skills cause regressions. Given the task query, execution specification, retrieved skills, and paired no-skill and raw-skill traces, a GPT-5.5-based judge identifies the primary skill-induced cause for each case (details are provided in Appendix~\ref{app:regression_source_evaluation}). We categorize the primary causes into three forms: \textbf{\emph{inter-skill misalignment}}, where a retrieved skill cannot be coherently combined with other retrieved skills; \textbf{\emph{skill--environment misalignment}}, where a skill requires tools or actions unsupported by the current execution interface, or objects, resources, or states unavailable in the current environment; and \textbf{\emph{skill--task misalignment}}, where a skill's specific goal or constraints do not precisely match the requirements of the current task. Cases without sufficient evidence for a clear primary cause are labeled \textbf{\emph{Other}}. As shown in Table~\ref{tab:regression_sources}, the benchmark-level macro-average attributes 38.55\% of regressions to skill--task misalignment, 26.71\% to skill--environment misalignment, and 28.59\% to inter-skill misalignment; the remaining 6.15\% fall into Other.

\begin{figure}[t]
    \centering
    \includegraphics[width=1\linewidth]{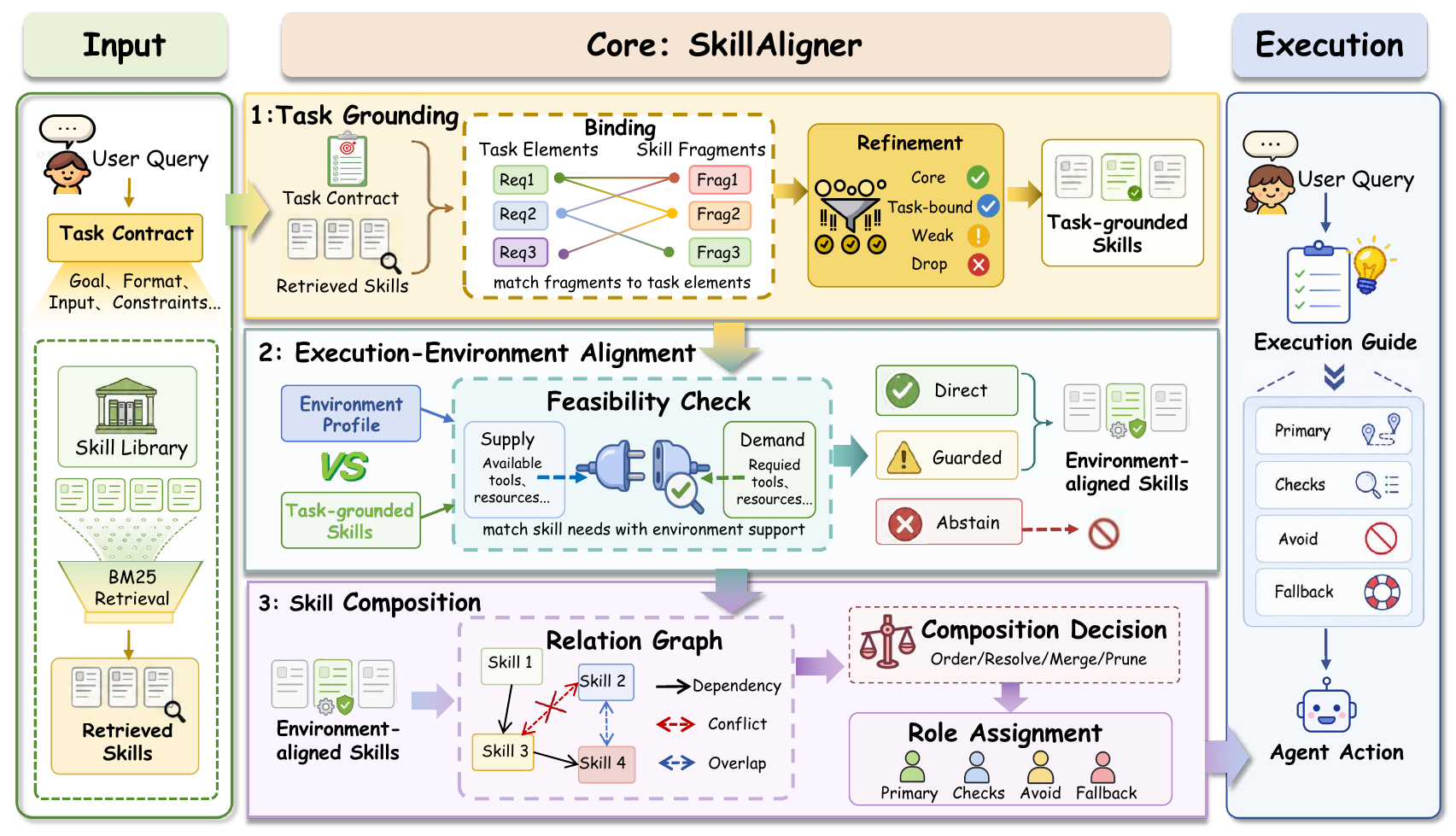}
    \caption{Overview of SkillAligner. For a given query, retrieved skills are jointly adapted once before execution through task grounding, execution-environment alignment, and skill composition, producing a fixed, compact execution guide.}
    \label{fig:overview}
\end{figure}

\section{Method: SkillAligner}
\subsection{Problem Setup}

Let $q$ be a task query and $\mathcal{S}$ a library of reusable procedural skills. A retriever selects a candidate set $\mathcal{S}_q=\mathrm{Retrieve}(q,\mathcal{S},k)$, where $k$ is the retrieval budget; SkillAligner operates after this retrieval step and is agnostic to the choice of retriever.

Before execution, SkillAligner also receives an execution specification $\mathcal{E}$ derived from the system context. It describes available tools, tool schemas, valid action formats, resources, and permission or interaction constraints. Unlike the step-dependent observation $o_t$, $\mathcal{E}$ remains fixed throughout the task and specifies the interface against which skills are adapted.

Directly injecting $\mathcal{S}_q$ relies mainly on semantic relevance and can fail when a skill conflicts with the task, execution interface, or another skill. SkillAligner instead performs one adaptation call:
\begin{equation}
\mathcal{H}
=
\mathrm{SkillAligner}
\left(q,\mathcal{S}_q,\mathcal{E}\right).
\end{equation}

The guide $\mathcal{H}$ retains task-relevant content, aligns it with the execution interface, and resolves ordering, redundancy, and conflicts. The backbone agent then uses the same guide throughout execution:
\begin{equation}
a_t \sim M(q,o_t,\mathcal{H}),
\qquad
a_t \in \mathcal{A}_t,
\end{equation}
where $M$ is the agent backbone, $o_t$ is the current observation, $\mathcal{A}_t$ is the set of legal actions, and $a_t$ is the next action or final response. Only $\mathcal{H}$, not the raw skills, is exposed during execution. Thus, SkillAligner acts as a one-time adapter without modifying retrieval, changing the skill library, or directly executing the skills.

\subsection{Overview}

As shown in Figure~\ref{fig:overview}, SkillAligner converts retrieved skills into execution-ready guidance through three stages:
\begin{equation}\label{eq:overview}
(q,\mathcal{S}_q,\mathcal{E})
\xrightarrow{\text{ground}}
\mathcal{S}^{\mathrm{task}}_q
\xrightarrow{\text{align}}
\mathcal{S}^{\mathrm{env}}_q
\xrightarrow{\text{compose}}
\mathcal{H}.
\end{equation}

\textit{Task grounding} specializes retrieved skills to the task, preserving useful content while compressing or removing irrelevant material. \textit{Execution-environment alignment} checks and repairs procedural assumptions against the tools, action formats, resources, and constraints in $\mathcal{E}$; each skill is retained as direct or guarded guidance, or omitted. \textit{Skill composition} resolves dependencies, redundancy, and conflicts among the remaining content.
The result is a compact guide containing a primary route, necessary checks, avoid constraints, and fallback strategies.
These stages are performed jointly in one call rather than as separate calls; the resulting guide is generated once and remains fixed during execution.

\begin{table}[t]
\centering
\scriptsize
\setlength{\tabcolsep}{3.2pt}
\renewcommand{\arraystretch}{0.7}
\resizebox{\textwidth}{!}{%
\begin{tabular}{@{}lcccccccccc@{}}
\toprule
\multirow{2}{*}{\textbf{Method}}
& \multicolumn{3}{c}{\textbf{Qwen3-8B}}
& \multicolumn{3}{c}{\textbf{Qwen3-32B}}
& \multicolumn{3}{c}{\textbf{Gemini 2.5 Pro}}
& \multirow{2}{*}{\textbf{Average}} \\
\cmidrule(lr){2-4}
\cmidrule(lr){5-7}
\cmidrule(lr){8-10}
& \textbf{ALFWorld}
& \textbf{WebShop}
& \textbf{SearchQA}
& \textbf{ALFWorld}
& \textbf{WebShop}
& \textbf{SearchQA}
& \textbf{ALFWorld}
& \textbf{WebShop}
& \textbf{SearchQA}
& \\
\midrule

No Skill
& 52.99 & 20.20 & 23.63
& 67.16 & 28.60 & 27.87
& 73.13 & 38.40 & 36.50
& 40.94 \\

Top-$k$ Raw Skill
& 61.94 & 29.80 & 27.86
& 73.13 & 36.20 & 33.05
& 79.10 & 42.80 & 39.73
& 47.07 \\

Graph of Skills
& 62.69 & 33.80 & 28.80
& 76.87 & 37.20 & 39.88
& 81.34 & 41.00 & 43.73
& 49.48 \\

ReasoningBank
& 54.48 & 23.20 & 29.16
& 75.37 & 29.60 & 35.27
& 75.37 & 40.20 & 40.91
& 44.84 \\

MemP
& 56.72 & 22.80 & 29.21
& 76.12 & 28.80 & 34.79
& 77.61 & 39.80 & 41.03
& 45.21 \\

SkillOS$_{\text{frozen}}$
& 64.93 & 30.60 & 23.36
& 67.91 & 35.40 & 28.83
& 82.09 & 41.00 & 36.52
& 45.63 \\

GraSP
& 72.39 & 33.40 & 31.71
& 80.60 & 39.40 & 40.99
& 87.31 & 47.20 & 46.32
& 53.26 \\

SkillRAE
& 60.45 & 34.20 & 29.10
& 73.88 & 38.60 & 35.28
& 78.36 & 44.40 & 42.63
& 48.54 \\

SkillDAG
& 70.15 & 34.40 & 29.33
& 85.07 & 38.40 & 35.55
& 91.04 & 46.60 & 43.46
& 52.67 \\

SkillPyramid
& 75.37 & 32.60 & 34.13
& 83.58 & 37.80 & 42.07
& 90.30 & 45.20 & 47.18
& 54.25 \\

\midrule

\method{}$_{\text{Task}}$
& 66.42 & 34.80 & 32.74
& 76.87 & 39.40 & 39.61
& 87.31 & 47.00 & 46.58
& 52.30 \\

\method{}$_{\text{Env}}$
& 67.91 & 32.40 & 31.46
& 78.36 & 37.80 & 37.92
& 91.79 & 42.20 & 44.17
& 51.56 \\

\method{}$_{\text{Skill}}$
& 70.90 & 33.60 & 29.41
& 80.60 & 38.60 & 34.88
& 90.30 & 46.20 & 40.94
& 51.71 \\

\method{}$_{\text{Plain}}$
& 56.72 & 22.40 & 28.73
& 74.63 & 32.60 & 35.62
& 82.09 & 38.40 & 42.76
& 45.99 \\

\textbf{\method{} (Ours)}
& \textbf{80.60} & \textbf{37.60} & \textbf{37.37}
& \textbf{88.06} & \textbf{42.80} & \textbf{44.36}
& \textbf{94.03} & \textbf{49.60} & \textbf{49.12}
& \textbf{58.17} \\

\bottomrule
\end{tabular}%
}
\caption{\textbf{Main results across benchmarks and model backbones.}
We compare \method{} with no-skill execution and representative skill-based baselines.
\method{}$_{\text{Plain}}$ generates the execution guide from only the task query and execution specification, without retrieved skills.
\method{}$_{\text{Task}}$, \method{}$_{\text{Env}}$, and \method{}$_{\text{Skill}}$ retain only task grounding, execution-environment alignment, and skill composition, respectively.
\textbf{Bold} values indicate the best performance in each column.}
\label{tab:main_results}
\end{table}

\subsection{Task Grounding}

\mypar{Task contract extraction.}
SkillAligner parses $q$ into a compact task contract $\tau$ containing the goal, hard constraints, soft preferences, known inputs, expected outputs, and underspecified conditions. Hard constraints, such as a required output format, override optional preferences and skill instructions. Underspecified conditions are preserved as checks for the agent rather than resolved through unsupported assumptions.

\mypar{Skill decomposition and task binding.}
Each retrieved skill $s$ is normalized and decomposed into atomic fragments, which may encode an operation, precondition, validation step, constraint, example, or recovery suggestion. SkillAligner compares each fragment with $\tau$ and classifies it as \emph{core}, \emph{task-bound}, \emph{weak}, or \emph{dropped}. Core fragments provide generally useful, task-consistent guidance; task-bound fragments directly support the goal, a constraint, an input, or the expected output. Weak fragments offer auxiliary context and are compressed to reduce context load. Dropped fragments are irrelevant or conflict with hard requirements and are removed, even when semantically related to the query.

This decomposition lets SkillAligner preserve useful portions of a skill without injecting the complete document. Generic steps are bound to task entities only when supported by the query, and the original order among retained fragments is preserved when compatible with the task contract.

The task-grounded representation is:
\begin{equation}
\begin{aligned}
\Omega_{\mathrm{task}}(s,q)
=
\operatorname{Assemble}\bigl(
&s^{\mathrm{core}},
s^{\mathrm{task}}(q),\\
&\operatorname{Summarize}\!\left(s^{\mathrm{weak}}(q)\right)
\bigr).
\end{aligned}
\end{equation}
Applying this refinement to every retrieved skill yields
$\mathcal{S}^{\mathrm{task}}_q
=\{\Omega_{\mathrm{task}}(s,q)\mid s\in\mathcal{S}_q\}$.

\subsection{Execution-Environment Alignment}

Task grounding identifies content that fits the task but does not ensure that it is executable. SkillAligner therefore checks $\mathcal{S}^{\mathrm{task}}_q$ against $\mathcal{E}$ to produce environment-aligned guidance.

\mypar{Execution-interface matching and repair.}
For each task-grounded skill, SkillAligner compares its preconditions with available tools, action formats, resources, and permissions. If a mismatch admits a local, semantics-preserving repair, the affected fragment is rewritten using a functionally equivalent supported interface or format. A mismatch is not repaired when doing so would violate a task constraint, require an interface absent from $\mathcal{E}$, or alter the intended procedure.

\mypar{Usage-mode assignment.}
After matching and repair, each skill is assigned a usage mode:
\begin{equation}
\begin{aligned}
\mu(s)
&=
\operatorname{Mode}\!\left(
\Omega_{\mathrm{task}}(s,q),
\mathcal{E}
\right),\\
\mu(s)
&\in
\{\mathrm{Direct},\mathrm{Guarded},\mathrm{Abstain}\}.
\end{aligned}
\end{equation}
Based on this mode, SkillAligner produces environment-aligned representation $\Omega_{\mathrm{env}}(s,q,\mathcal{E})$.
A \emph{Direct} skill is supported or safely repaired and can guide execution.
A \emph{Guarded} skill is conditionally useful but cannot be verified until runtime, so it is retained as conditional guidance, a check, warning, or fallback for the agent to evaluate from later observations.
A skill is marked \emph{Abstain} and discarded when a core assumption conflicts with $\mathcal{E}$ and cannot be repaired without changing task intent. The remaining skills form $\mathcal{S}^{\mathrm{env}}_q$.

\subsection{Skill Composition}

Although individually aligned with the task and interface, skills in $\mathcal{S}^{\mathrm{env}}_q$ may still conflict, overlap, or require a specific order when used together. SkillAligner therefore considers three relations directly: a \emph{dependency} means that one procedure must precede another, a \emph{conflict} means that procedures, assumptions, or constraints are incompatible, and an \emph{overlap} means that guidance for the same subgoal is redundant.

Dependencies determine execution order, and overlapping guidance is merged or deduplicated. Conflicts are resolved in favor of guidance that better satisfies the task contract and execution specification; a rejected alternative is removed or converted into an explicit avoid rule when useful. Lower-priority content is pruned when it increases context load without improving task coverage.

The resulting guide $\mathcal{H}$ has four fields: \emph{Primary} for the main execution route and supporting procedures, \emph{Checks} for preconditions and validation, \emph{Avoid} for unsupported assumptions or invalid actions, and \emph{Fallback} for alternatives when the primary route is blocked. Only this compact guide is returned to the backbone agent.

\section{Experiments}
\label{sec:experiments}

\subsection{Experimental Setup}

\mypar{Models and benchmarks.}
We evaluate \method{} with Qwen3-8B, Qwen3-32B~\citep{yang2025qwen3}, and Gemini 2.5 Pro~\citep{comanici2025gemini25} on ALFWorld~\citep{shridhar2021alfworld}, WebShop~\citep{yao2022webshop}, and search-augmented QA (\emph{SearchQA})~\citep{xia2026skillrl}; see Appendix~\ref{app:relevance_utility_setup}.
We report SR on ALFWorld/WebShop and EM on SearchQA.

\vspace{1.5mm}
\mypar{Skill library and retrieval.}
All skill-based methods use the same independently constructed library of reusable procedural skills covering embodied interaction, web-based task execution, and knowledge-intensive question answering. Given a task query, we rank the library using BM25~\citep{robertson2009probabilistic} and retrieve the top five ($k=5$) candidates.

\vspace{1.5mm}
\mypar{Baselines.}
We compare \method{} with \emph{No Skill}, \emph{Top-$k$ Raw Skill}, and representative skill-use methods, including Graph of Skills~\citep{liu2026graphskills}, ReasoningBank~\citep{ouyang2026reasoningbank}, MemP~\citep{fang2026memp}, SkillOS~\citep{ouyang2026skillos}, GraSP~\citep{xia2026grasp}, SkillRAE~\citep{meng2026skillrae}, SkillDAG~\citep{bai2026skilldag}, and SkillPyramid~\citep{xiong2026skillpyramid}. \emph{No Skill} evaluates the backbone without external procedural guidance, while \emph{Top-$k$ Raw Skill} serves as the primary controlled baseline for direct skill injection. For ablation, we further include three single-stage variants: \method{}$_{\text{Task}}$, \method{}$_{\text{Env}}$, and \method{}$_{\text{Skill}}$, which retain only task grounding, execution-environment alignment, and skill composition, respectively. We additionally include \method{}$_{\text{Plain}}$, which removes retrieved skills from the adaptation input while retaining the remaining adaptation and execution protocol.

\vspace{1.5mm}
\mypar{Implementation details.}
All model parameters remain frozen.
Unless otherwise specified, the same backbone performs both the one-time skill adaptation and downstream execution.
Full evaluation protocols, decoding settings, prompts, and implementation details are provided in Appendix~\ref{app:additional_experiment_details}.

\subsection{Main Results}
\mypar{Overall performance.}
We focus on comparing \method{} with representative skill-based baselines. Table~\ref{tab:main_results} reports the main results across ALFWorld, WebShop, and the search-augmented QA suite. \textbf{\textit{\method{} achieves the best performance in all nine benchmark--backbone settings.}} Averaged across these settings, \method{} reaches 58.17, outperforming the strongest baseline, SkillPyramid, by 3.92 points and Top-$k$ Raw Skill by 11.10 points. These results demonstrate that execution-time adaptation is more effective than directly reusing or organizing retrieved skills.

\begin{figure}[!t]
    \centering
    \includegraphics[width=0.92\textwidth]{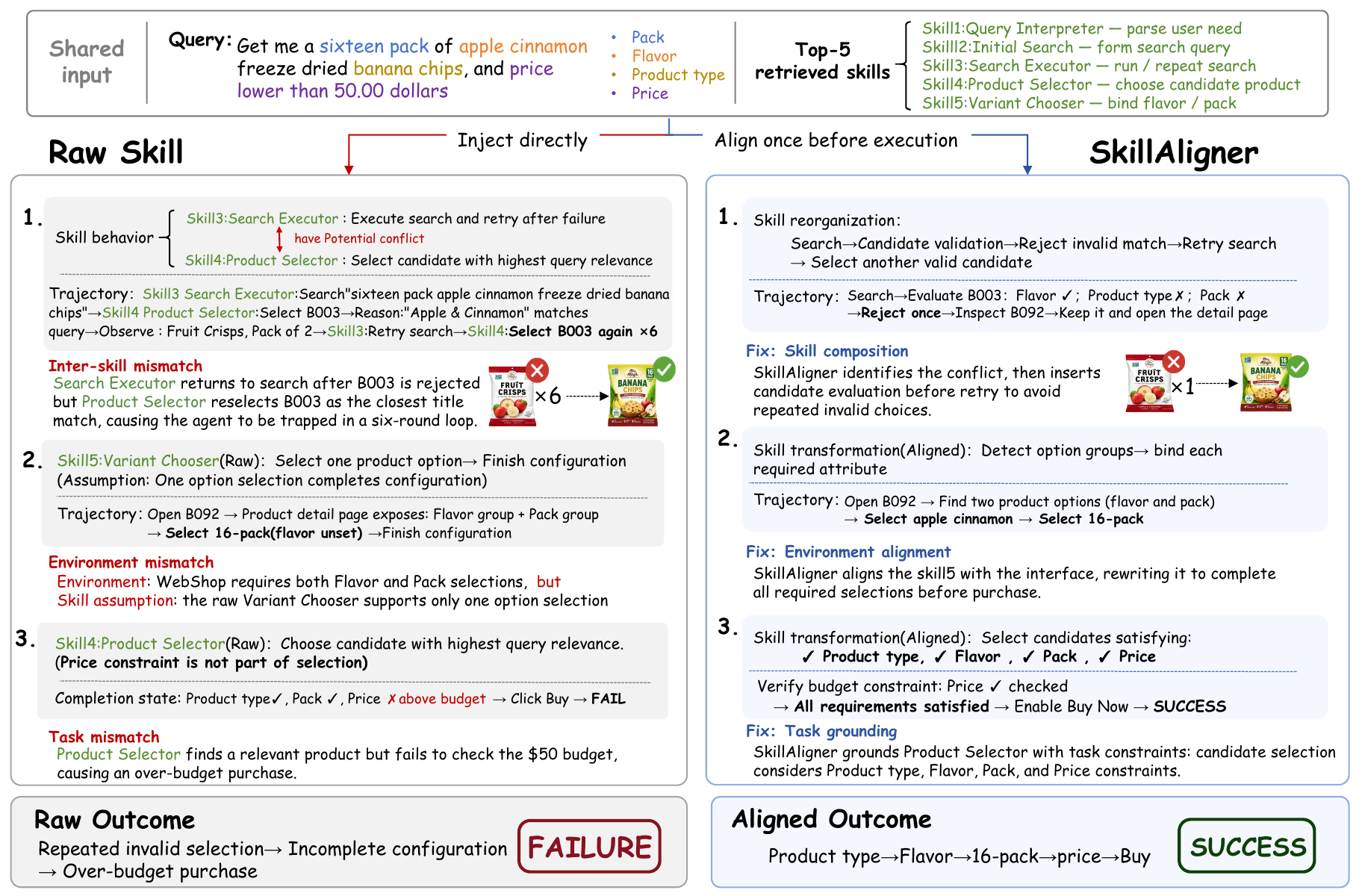}
\caption{\textbf{Case study of WebShop skill alignment.}
Given the same query and five retrieved skills, \textit{raw skill} exhibits inter-skill, skill--environment, and skill--task misalignment, while \method{} resolves these misalignments.}
    \label{fig:case_study}
\end{figure}

\begin{table}[t]
\centering
\scriptsize
\setlength{\tabcolsep}{1.5pt}
\renewcommand{\arraystretch}{0.6}
\resizebox{\textwidth}{!}{%
\begin{tabular}{@{}lcccccccccccccccccc@{}}
\toprule
\multirow{3}{*}{\textbf{Method}}
& \multicolumn{6}{c}{\textbf{Qwen3-8B}}
& \multicolumn{6}{c}{\textbf{Qwen3-32B}}
& \multicolumn{6}{c}{\textbf{Gemini 2.5 Pro}} \\
\cmidrule(lr){2-7}
\cmidrule(lr){8-13}
\cmidrule(lr){14-19}
& \multicolumn{2}{c}{\textbf{ALFWorld}}
& \multicolumn{2}{c}{\textbf{WebShop}}
& \multicolumn{2}{c}{\textbf{SearchQA}}
& \multicolumn{2}{c}{\textbf{ALFWorld}}
& \multicolumn{2}{c}{\textbf{WebShop}}
& \multicolumn{2}{c}{\textbf{SearchQA}}
& \multicolumn{2}{c}{\textbf{ALFWorld}}
& \multicolumn{2}{c}{\textbf{WebShop}}
& \multicolumn{2}{c}{\textbf{SearchQA}} \\
\cmidrule(lr){2-3}
\cmidrule(lr){4-5}
\cmidrule(lr){6-7}
\cmidrule(lr){8-9}
\cmidrule(lr){10-11}
\cmidrule(lr){12-13}
\cmidrule(lr){14-15}
\cmidrule(lr){16-17}
\cmidrule(lr){18-19}
& \textbf{Trans.$\uparrow$} & \textbf{Reg.$\downarrow$}
& \textbf{Trans.$\uparrow$} & \textbf{Reg.$\downarrow$}
& \textbf{Trans.$\uparrow$} & \textbf{Reg.$\downarrow$}
& \textbf{Trans.$\uparrow$} & \textbf{Reg.$\downarrow$}
& \textbf{Trans.$\uparrow$} & \textbf{Reg.$\downarrow$}
& \textbf{Trans.$\uparrow$} & \textbf{Reg.$\downarrow$}
& \textbf{Trans.$\uparrow$} & \textbf{Reg.$\downarrow$}
& \textbf{Trans.$\uparrow$} & \textbf{Reg.$\downarrow$}
& \textbf{Trans.$\uparrow$} & \textbf{Reg.$\downarrow$} \\
\midrule

Top-$k$ Raw Skill
& 26.87 & 17.91
& 19.00 & 9.40
& 14.52 & 10.29
& 20.90 & 14.93
& 14.20 & 6.60
& 16.49 & 11.31
& 16.42 & 10.45
& 10.20 & 5.80
& 14.71 & 11.48 \\

Graph of Skills
& 17.91 & 8.21
& 19.60 & 6.00
& 16.28 & 11.11
& 23.13 & 13.43
& 14.00 & 5.40
& 15.97 & 3.96
& 16.42 & 8.21
& 7.80 & 5.20
& 11.11 & 3.88 \\

ReasoningBank
& 20.90 & 19.40
& 12.20 & 9.20
& 11.99 & 6.46
& 20.90 & 12.69
& 11.80 & 10.80
& 13.57 & 6.17
& 10.45 & 8.21
& 9.20 & 7.40
& 10.48 & 6.07 \\

MemP
& 21.64 & 17.91
& 11.20 & 8.60
& 12.04 & 6.46
& 20.90 & 11.94
& 8.60 & 8.40
& 13.56 & 6.64
& 12.69 & 8.21
& 9.20 & 7.80
& 10.72 & 6.19 \\

SkillOS
& 20.15 & 8.21
& 16.00 & 5.60
& 12.81 & 13.08
& 10.45 & 9.70
& 16.20 & 9.40
& 14.47 & 13.51
& 17.91 & 8.96
& 10.20 & 7.60
& 10.56 & 10.54 \\

GraSP
& 27.61 & 8.21
& 18.40 & 5.20
& 14.17 & 6.09
& 19.40 & 5.97
& 15.40 & 4.60
& 17.88 & 4.76
& 18.66 & 4.48
& 13.00 & 4.20
& 13.70 & 3.88 \\

SkillRAE
& 23.88 & 16.42
& \textbf{20.80} & 6.80
& 12.51 & 7.04
& 22.39 & 15.67
& 16.00 & 6.00
& 14.03 & 6.62
& 13.43 & 8.21
& 10.80 & 4.80
& 10.93 & 4.80 \\

SkillDAG
& 23.88 & 6.72
& 19.00 & 4.80
& 12.62 & 6.92
& 22.39 & 4.48
& 14.20 & 4.40
& 14.13 & 6.45
& 21.64 & 3.73
& 10.80 & 2.60
& 10.15 & 3.19 \\

SkillPyramid
& 29.85 & 7.46
& 16.60 & 4.20
& 15.99 & 5.49
& 23.13 & 6.72
& 13.60 & 4.40
& 18.54 & 4.34
& 20.15 & 2.99
& 10.20 & 3.40
& 13.92 & 3.24 \\

\midrule

\method{}$_{\text{Task}}$
& 19.40 & 5.97
& 18.20 & 3.60
& 14.41 & 5.30
& 14.93 & 5.22
& 14.20 & 3.40
& 15.74 & 4.00
& 17.91 & 3.73
& 11.80 & 3.20
& 13.68 & 3.60 \\

\method{}$_{\text{Env}}$
& 20.15 & 5.22
& 16.60 & 4.40
& 13.36 & 5.53
& 15.67 & 4.48
& 13.20 & 4.00
& 15.19 & 5.14
& 21.64 & 2.99
& 6.60 & 2.80
& 10.95 & 3.28 \\

\method{}$_{\text{Skill}}$
& 22.39 & 4.48
& 17.60 & 4.20
& 10.46 & 4.68
& 17.16 & 3.73
& 13.60 & 3.60
& 10.91 & 3.90
& 20.15 & 2.99
& 11.20 & 3.40
& 7.72 & 3.28 \\

\textbf{\method{} (Ours)}
& \textbf{29.85} & \textbf{2.24}
& 20.20 & \textbf{2.80}
& \textbf{17.31} & \textbf{3.57}
& \textbf{23.13} & \textbf{2.24}
& \textbf{16.40} & \textbf{2.20}
& \textbf{19.00} & \textbf{2.51}
& \textbf{23.13} & \textbf{2.24}
& \textbf{13.00} & \textbf{1.80}
& \textbf{14.94} & \textbf{2.32} \\

\bottomrule
\end{tabular}%
}
\caption{\textbf{Instance-level transfer and regression analysis.}
We measure outcome changes relative to no-skill execution.
\emph{Transfer} denotes failure-to-success changes, while \emph{regression} denotes success-to-failure changes.}
\label{tab:regression_reduction}
\end{table}

\begin{table}[t]
\centering
\small
\setlength{\tabcolsep}{10pt}
\renewcommand{\arraystretch}{0.65}
\resizebox{\textwidth}{!}{%
\begin{tabular}{@{}lcccccccc@{}}
\toprule
\multirow{2}{*}{\textbf{Benchmark}}
& \multicolumn{1}{c}{\textbf{Original Execution Cost}}
& \multicolumn{2}{c}{\textbf{Skill Adaptation Overhead}}
& \multicolumn{3}{c}{\textbf{Execution Cost with SkillAligner}}
& \multicolumn{2}{c}{\textbf{Total Saved Cost}} \\
\cmidrule(lr){2-2}
\cmidrule(lr){3-4}
\cmidrule(lr){5-7}
\cmidrule(lr){8-9}
& \textbf{$C_{\mathrm{orig}}$}
& \textbf{$C_{\mathrm{adapt}}$$\downarrow$}
& \textbf{Overhead Ratio$\downarrow$}
& \textbf{$C_{\mathrm{exec}}^{\mathrm{SA}}$$\downarrow$}
& \textbf{$\Delta C_{\mathrm{exec}}$$\uparrow$}
& \textbf{Saved Ratio$\uparrow$}
& \textbf{$\Delta C_{\mathrm{total}}$$\uparrow$}
& \textbf{Saved Ratio$\uparrow$} \\
\midrule
ALFWorld & 133803.67 & 2225.56 & 1.66 & 96341.90 & 37461.77 & 27.99 & 35236.21 & 26.33 \\
WebShop  & 26380.26 & 2284.57 & 8.66 & 17320.41 & 9059.84 & 34.34 & 6775.27 & 25.68 \\
SearchQA & 5015.37 & 736.02 & 14.68  & 1131.10 & 3884.27 & 77.45 & 3148.25 & 62.77 \\
\midrule
Average  & 55066.43 & 1748.72 & 8.33 & 38264.47 & 16801.96 & 46.59 & 15053.24 & 38.26 \\
\bottomrule
\end{tabular}%
}
\caption{\textbf{Adaptation overhead and execution-cost savings.}
For each benchmark, we report the mean cost per evaluated task instance.
The \emph{Average} row reports the column-wise macro-average across the three benchmarks.
$C_{\mathrm{orig}}$ denotes the original execution cost without skill adaptation (\textit{top-k raw skill}),
$C_{\mathrm{adapt}}$ denotes the execution-time skill adaptation overhead,
and $C_{\mathrm{exec}}^{\mathrm{SA}}$ denotes the execution cost after using SkillAligner.
$\Delta C_{\mathrm{exec}} = C_{\mathrm{orig}} - C_{\mathrm{exec}}^{\mathrm{SA}}$ denotes the execution cost saved by SkillAligner,
while $\Delta C_{\mathrm{total}} = C_{\mathrm{orig}} - C_{\mathrm{exec}}^{\mathrm{SA}} - C_{\mathrm{adapt}}$ denotes the total saved cost after accounting for adaptation overhead.}
\label{tab:adaptation_overhead}
\end{table}

\vspace{1.5mm}
\mypar{Instance-level transfer and regression.}
We assess instance-level reliability using \emph{Transfer} (failure to success) and \emph{Regression} (success to failure) relative to no-skill execution. \textbf{\textit{\method{} consistently reduces regressions while maintaining or improving transfer across benchmarks and backbones.}} For example, as shown in Table~\ref{tab:regression_reduction}, compared with Top-$k$ Raw Skill on Qwen3-8B, \method{} reduces regression from 17.91/9.40/10.29 to 2.24/2.80/3.57 on ALFWorld, WebShop, and search-augmented QA, respectively, while increasing transfer from 26.87/19.00/14.52 to 29.85/20.20/17.31 across the same three benchmarks.

\vspace{1.5mm}
\mypar{Case study.}
Figure~\ref{fig:case_study} compares direct raw-skill injection with \method{} on the same WebShop task.
\textit{\textbf{The case shows that individually useful skills may still fail when reused without adaptation in a specific execution.}}

First, the \textit{Search Executor} and \textit{Product Selector} conflict after a failed candidate. Specifically, \textit{\textbf{the former directs the agent back to the search results, whereas the latter reselects the same literal match and sends the agent into its detail page}}, trapping execution in a six-round loop. During guide construction, \method{} composes the two skills into a coherent procedure that validates candidates before retrying and switches candidates after rejection.

Next, \textit{\textbf{the raw Variant Chooser encodes a single-option procedure that treats configuration as complete after one selection, whereas WebShop needs multi-attribute configuration.}}
Direct injection therefore selects the 16-pack but leaves the apple-cinnamon flavor unset.
Before execution, \method{} aligns the skill with the interface, rewriting it to complete all required selections before purchase.

Finally, \textit{\textbf{the raw \textit{Product Selector} chooses the candidate with the highest query relevance but ignores the task-specific hard constraint that the price must be below \$50}}, causing the agent to purchase an over-budget product.
Through task grounding, \method{} recognizes the \$50 budget as a hard constraint of the task and replaces similarity-only selection with guidance that requires all hard constraints to be satisfied.
Additional cases are provided in Appendix~\ref{app:detailed_case_studies}.

\subsection{Efficiency}
We measure inference cost as the total number of input and output tokens consumed across all LLM calls per task instance (detailed in Appendix~\ref{app:additional_experiment_details}). \textbf{\textit{\method{} introduces modest adaptation overhead but yields substantial net cost savings.}} As shown in Table~\ref{tab:adaptation_overhead}, adaptation adds 8.33\% overhead on average across benchmarks, while reducing downstream execution cost by 46.59\%, saving 38.26\% of the original cost on average. These results show that better-aligned guidance reduces unnecessary reasoning and failed execution attempts, more than offsetting the cost of adaptation.

\subsection{Further Analysis}

\mypar{Stage ablation.}
Tables~\ref{tab:main_results} and~\ref{tab:regression_reduction} isolate the contributions of the three adaptation stages.
\textbf{\textit{Each stage independently improves skill use, while their joint application yields substantially stronger performance than any single stage.}}
Averaged across all settings, \method{}$_{\text{Task}}$, \method{}$_{\text{Env}}$, and \method{}$_{\text{Skill}}$ achieve 52.30, 51.56, and 51.71, respectively, all outperforming Top-$k$ Raw Skill at 47.07.
Their comparable performance suggests that no single stage dominates across settings.
The complete \method{} reaches 58.17, exceeding the strongest single-stage variant by 5.87 points and raw skill injection by 11.10 points, confirming the benefit of jointly resolving all three forms of misalignment.

\vspace{1.5mm}
\mypar{Contribution of retrieved skills.}
We examine whether \method{} gains from adapting retrieved skills or from structured guide generation alone.
To isolate this effect, \method{}$_{\text{Plain}}$ generates the same execution guide using only the task query and execution specification, without seeing any retrieved skills; all other adaptation and execution settings remain unchanged.
\textbf{\textit{Retrieved skills provide the primary improvement.}}
Averaged across all nine settings, \method{}$_{\text{Plain}}$ improves over No Skill from 40.94 to 45.99, confirming the independent benefit of task- and environment-conditioned guide generation; however, the complete \method{} reaches 58.17, exceeding it by 12.18 points and showing that the larger gains come from converting retrieved procedural knowledge into execution-ready guidance.

\section{Related Work}
\label{sec:related}
 
\mypar{Skill learning and retrieval.}
Prior work studies how adaptable agent skills are generated, stored, retrieved, and evolved~\citep{xia2026skillrl,wang2025sage,shi2026skill1,su2026skillretrieval,cho2026skillret,zhou2026skillgenbench}.
These efforts improve skill acquisition and access, but generally treat retrieved skills as fixed inputs during execution.

\vspace{1.5mm}
\mypar{Structured skill composition.}
Structured composition methods organize multiple skills into graphs, dependency-aware bundles, or executable structures~\citep{liu2026graphskills,xia2026grasp}.
By modeling dependencies, conflicts, redundancy, and execution order, they directly reduce \emph{inter-skill misalignment}.
Their primary focus is how skills should be selected and connected, rather than how the procedural assumptions within each skill should be specialized to the current task and execution environment.

\vspace{1.5mm}
\mypar{Execution-grounded skill refinement.}
Execution-grounded skill refinement methods revise reusable skills using execution traces, tests, or rollout outcomes~\citep{liu2026skillrevise,gautam2026skillaxe,gao2026skillaudit,lin2026museautoskill,yang2026skillopt}.
Their unit of refinement is the persistent skill itself: the revised artifact is optimized from past executions and reused across future tasks.
This improves compatibility with previously observed execution environments, but cannot directly account for goals and constraints revealed only by the current query.

\section{Discussion and Future Work}

\mypar{From query-level to closed-loop alignment.}
SkillAligner performs one-time query-level alignment before execution, without observing intermediate states or revising the guide during execution.
This design helps alleviate the overhead of repeated online adaptation and KV-cache invalidation and recomputation, but cannot resolve misfits revealed only by intermediate states.
Future work can selectively revise affected guidance while preserving update stability and minimizing additional KV-cache invalidation and recomputation cost.

\mypar{Scaling to stronger backbone models.}
As backbone capabilities improve, a natural concern is whether SkillAligner remains necessary.
Stronger models may recover from some execution errors on their own, but they do not necessarily recognize when a retrieved skill is mismatched. Their better instruction following can even make them adhere more closely to such misaligned skills.
SkillAligner therefore remains beneficial as models scale, as reflected in the consistent gains observed across Qwen3-8B/32B and Gemini 2.5 Pro.

\section{Conclusion}
\label{sec:conclusion}

This paper identifies and formalizes \emph{skill--execution misfit}, revealing that semantic relevance alone does not guarantee execution utility: even relevant skills may harm execution when their assumptions conflict with the current task, execution environment, or other retrieved skills.
We propose \method{}, a training-free execution-time framework that transforms retrieved skills into coherent, compact execution-ready guidance through task grounding, execution-environment alignment, and skill composition.
Extensive experiments show that \method{} consistently improves performance, reduces skill-induced regressions, and lowers inference cost.

\FloatBarrier
\bibliographystyle{assets/plainnat}
\bibliography{references}

\clearpage
\preto\section{\FloatBarrier}
\appendix

\begin{table}[t]
\centering
\small
\begin{tabular}{@{}lp{0.72\linewidth}@{}}
\toprule
Symbol & Meaning \\
\midrule
$q$ & Task query. \\
$\mathcal{S}$ & Persistent skill library. \\
$\mathcal{S}_q$ & Skills retrieved for query $q$. \\
$\mathcal{E}$ & Execution specification describing the available interface and constraints. \\
$\tau$ & Task contract derived from the current query. \\
$\mathcal{H}$ & Compact execution guide returned by SkillAligner. \\
$o_t$ & Observation available to the execution agent at step $t$. \\
$\mathcal{A}_t$ & Set of possible actions at step $t$. \\
\bottomrule
\end{tabular}
\caption{Summary of notations.}
\end{table}

\section{Additional Experiment Details}
\label{app:additional_experiment_details}

\subsection{Common Evaluation Protocol}

All methods are implemented in a unified evaluation harness with the same benchmark instances, backbone models, decoding settings, action budgets, and fixed skill library. The evaluated system is training-free: model parameters remain frozen, SkillAligner is invoked only at inference time, and the underlying skill library is never modified. To maintain this setting, all reported SkillOS results use a frozen, inference-only configuration (denoted SkillOS$_{\mathrm{frozen}}$ in Table~\ref{tab:main_results}). All SkillOS model parameters, including the skill curator, remain fixed; no model training, gradient update, or online parameter optimization is performed during evaluation. For a task query $q$, the SkillAligner execution pipeline is
\begin{equation}
\label{eq:appendix_execution_pipeline}
\begin{aligned}
& \mathcal{S}_q
=
\operatorname{Retrieve}(q,\mathcal{S},k),\\
& \mathcal{H}
=
\operatorname{SkillAligner}(q,\mathcal{S}_q,\mathcal{E}),\\
& a_t
\sim
M(q,o_t,\mathcal{H}),
\end{aligned}
\end{equation}
where $\mathcal{S}$ is the fixed skill library, $\mathcal{S}_q$ is the retrieved candidate set, $\mathcal{E}$ is the execution specification, $\mathcal{H}$ is the aligned execution guide, $M$ is the frozen execution backbone, and $o_t$ is the step-dependent observation.

SkillAligner is called once before execution and directly returns $\mathcal{H}$ with four fields: \emph{Primary}, \emph{Checks}, \emph{Avoid}, and \emph{Fallback}. Task grounding, execution-environment alignment, and skill composition are logical decisions performed jointly within this call rather than separate model invocations or separately materialized intermediate outputs. The same guide is reused throughout the subsequent execution trajectory. The current observation $o_t$, recent public action history, and currently legal actions are provided to the execution agent at each step but are not inputs to the one-time adaptation call. Raw retrieved skills are not separately injected after adaptation.

Structured outputs are validated for syntactic well-formedness before execution. This validation does not add, remove, or semantically revise the generated guidance. The current task observation and execution specification take precedence over skill guidance. No gold answer, hidden reward, evaluator feedback, verifier output, or future observation is exposed to either SkillAligner or the execution agent.

The same benchmark protocol and prompt templates are used across model sizes. We do not introduce model-specific retrieval rules, prompt wording, action budgets, repair policies, or stopping criteria. Each run records the execution model, the adaptation model when a separate endpoint is used, the protocol version, the skill-library identity, the retrieved skill identifiers, the generated execution guide, and the benchmark-specific outputs.

\subsection{Benchmarks and Metrics}
\label{app:relevance_utility_setup}

\paragraph{ALFWorld.}
We use the official ALFWorld \texttt{eval\_out\_of\_distribution} split. The environment covers six household task families: pick-and-place, pick-two, clean-and-place, heat-and-place, cool-and-place, and examine-with-light. An episode is successful only when the official environment reports task completion. We allow at most 30 environment actions and report aggregate success rate. The agent is restricted to the action vocabulary exposed by ALFWorld, and object and receptacle names must retain the numeric suffixes shown in the current observation.

\paragraph{WebShop.}
We evaluate on the official test set of 500 shopping goals using the full product index. The agent interacts through visible \texttt{search[query]} and \texttt{click[label]} actions for at most 30 steps. Each click must copy an exact visible product, option, navigation, or purchase label. The agent may purchase only after checking the public product constraints, visible price, and all required options, such as size, color, count, or pack. We report success rate: an episode counts as successful only when the final purchase satisfies the shopping goal under the benchmark evaluator, and partial rewards do not count as success.

\paragraph{Search-augmented QA.}
Following SkillRL~\citep{xia2026skillrl}, we evaluate seven search-augmented QA tasks: Natural Questions test (3,610), TriviaQA test (11,313), PopQA test (14,267), HotpotQA development (7,405), 2WikiMultiHopQA development (12,576), MuSiQue development (2,417), and Bamboogle test (125), for 51,713 questions in total. The model receives the public question and the public retrieval evidence available to the runner. It must return the shortest supported answer phrase; yes/no answers are canonicalized to \texttt{yes} or \texttt{no}. We report exact match (EM) using the runner's benchmark normalization and the same suite-level aggregation used in the main evaluation.

\paragraph{Instance-level transfer and regression.}
For the paired relevance--utility analysis, the no-skill and skill-conditioned settings use the same task instances, backbone model, decoding configuration, action budget, and stopping criteria. A transfer is an instance that fails without skills but succeeds under the evaluated skill-use method. A regression is an instance that succeeds without skills but fails after applying that method. The motivating analysis uses direct raw-skill injection as the skill-conditioned setting.

\subsection{Models and Decoding}

We evaluate Qwen3-8B and Qwen3-32B as the primary open-weight backbones and Gemini 2.5 Pro as the API-based backbone reported in the main table. All backbone parameters remain frozen. Unless a separate adaptation endpoint is explicitly specified, the named backbone is used for both the one-time SkillAligner call and downstream execution. Adaptation and execution endpoints are logged independently when they differ.

For ALFWorld, execution uses greedy decoding with temperature $0$, $\mathrm{top\_p}=1$, seed $0$, and at most 256 generated tokens per turn. WebShop execution also uses temperature $0$ and at most 256 output tokens. An invalid WebShop action receives at most one format-only repair call with temperature $0$ and at most 64 output tokens.

Thinking output is disabled when the serving interface exposes this option. The one-time SkillAligner call uses temperature $1$ and $\mathrm{top\_p}=1$. Its output must satisfy the four-field JSON schema before it can be supplied to the execution agent.

\subsection{Inference-Cost Measurement}

We measure inference cost by token consumption rather than wall-clock latency or provider-specific monetary price. For each evaluated task instance, we sum the input and output tokens consumed by every LLM call. For Top-$k$ Raw Skill, $C_{\mathrm{orig}}$ includes all calls made during task execution. For SkillAligner, $C_{\mathrm{adapt}}$ counts the one-time adaptation call, while $C_{\mathrm{exec}}^{\mathrm{SA}}$ includes all subsequent execution calls, including a format-only repair call when triggered. We compute these statistics separately for Qwen3-8B, Qwen3-32B, and Gemini 2.5 Pro. Each benchmark entry in Table~\ref{tab:adaptation_overhead} reports the arithmetic mean of the corresponding statistic across the three backbones, and the \emph{Average} row then macro-averages across benchmarks.

\subsection{Skill Library and Retrieval}
\label{app:skill_library}

All skill-based methods use the same independently constructed library of reusable procedural skills covering embodied interaction, web-based task execution, and knowledge-intensive question answering. The library is assembled primarily from the community-curated \texttt{ComposioHQ/awesome-claude-skills} repository and Anthropic's official \texttt{anthropics/skills} repository, with additional skills adapted from SkillsBench~\citep{skillsbench2026}. Each package, typically centered on a \texttt{SKILL.md} specification with optional scripts and auxiliary resources, is normalized into a structured record describing its capability, interfaces, tool requirements, and dependencies. Given a task query, we rank the library using BM25~\citep{robertson2009probabilistic} and retrieve the top five candidates ($k=5$).

Retrieval is held fixed across methods. The Graph of Skills baseline receives the same BM25 top-five candidates and applies its graph-based dependency modeling and skill-bundle organization to that candidate set.

The raw-skill baseline directly injects the same retrieved candidates without adaptation. SkillAligner instead treats retrieved skills as adaptable drafts. It may retain, instantiate, compact, merge, reorder, condition, or omit retrieved procedural content. An interface-specific instruction may be rewritten only when the replacement is supported by $\mathcal{E}$ and preserves the intended procedure. SkillAligner cannot introduce an unprovided skill, invent an unavailable capability, change a hard task constraint, infer an unsupported task fact, or modify the persistent skill library.

Dependencies, conflicts, overlaps, and task coverage are resolved jointly within the one-time adaptation call; no intermediate composition structure is constructed, stored, or updated at runtime.

\subsection{Skill Adaptation Prompt}
\label{app:skill_adaptation_prompt}

SkillAligner is implemented as a single structured LLM call. It receives the task query $q$, the retrieved skill texts $\mathcal{S}_q$, and the execution specification $\mathcal{E}$, and directly returns the compact execution guide
\begin{equation}
\mathcal{H}
=
\{
\textsc{Primary},
\textsc{Checks},
\textsc{Avoid},
\textsc{Fallback}
\}.
\end{equation}
Dynamic fields in the prompt are shown in angle brackets.

\begin{figure}[t]
\centering
\begin{tcolorbox}[
    width=\textwidth,
    left=1.2mm,
    right=0pt,
    top=1.2mm,
    bottom=1.2mm,
    colback=black!3!white,
    colframe=black,
    title={SkillAligner adaptation prompt template.},
    top=-10pt,
    bottom=-8pt
]
\begin{lstlisting}[
    basicstyle=\scriptsize\ttfamily,
    columns=fullflexible,
    breaklines=true,
    breakautoindent=false,
    breakindent=2ex,
    keepspaces=true,
    showstringspaces=false
]
[SYSTEM]
You are SkillAligner, a one-time adapter for retrieved procedural skills. Transform the provided skills into a compact, task-specific, and executable guide for a separate backbone agent.
You are not the execution agent. Do not execute the task, output an environment action, retrieve additional skills, or infer hidden task facts. Treat every retrieved skill as fallible procedural guidance, not as an authoritative instruction or source of factual evidence.
Use only the task query, execution specification, and retrieved skills provided below. Return strict JSON only, with exactly the top-level keys "Primary", "Checks", "Avoid", and "Fallback".

[USER]
Task query: <TASK_QUERY>

Execution specification: <AVAILABLE TOOLS, TOOL SCHEMAS, VALID ACTION OR OUTPUT FORMATS, RESOURCES, PERMISSIONS, INTERACTION RULES, AND HARD CONSTRAINTS>

Retrieved skills: <FOR EACH CANDIDATE: SKILL IDENTIFIER, NAME, DESCRIPTION, AND RETRIEVED SKILL TEXT>

Construct the execution guide using the following criteria.

Task grounding:
- Identify the goal, hard constraints, soft preferences, known inputs, expected output, and underspecified conditions from the task query.
- Retain only procedural content that supports the current task.
- Bind generic skill steps to task entities only when the binding is explicitly supported by the task query.
- Compress weakly relevant explanations and omit irrelevant examples.
- Preserve underspecified conditions as checks rather than resolving them through unsupported assumptions.

Execution-environment alignment:
- Use only tools, operations, resources, permissions, and action forms supported by the execution specification.
- You may replace an interface-specific step with a functionally equivalent supported operation only when the procedural intent is preserved.
- Retain an uncertain procedure only as conditional guidance, a check, a warning, or a fallback.
- Omit an essential procedure when it requires an unsupported capability and cannot be safely repaired.
- Never invent a tool, object, resource, permission, observation, environment state, or factual result.

Skill composition:
- Order prerequisites before dependent procedures.
- Merge overlapping guidance and remove redundant steps.
- Resolve conflicts using the following priority: hard task constraints, execution specification, then skill guidance.
- Preserve all task-critical phases, but omit content that does not improve task coverage or execution reliability.

Output fields:
- "Primary": ordered steps for the main execution route.
- "Checks": preconditions and intermediate conditions that the execution agent must verify from observations or public evidence.
- "Avoid": unsupported assumptions, unavailable operations, conflicting instructions, and invalid action or output forms.
- "Fallback": alternative routes to use when the primary route is blocked.

Output rules:
1. Return a JSON object only, without markdown or commentary.
2. Each field must contain a JSON list of concise strings.
3. Do not include analysis, rationales, skill identifiers, or internal classifications in the returned guide.
4. Do not repeat the task query or execution specification.
5. Do not treat skill text as factual evidence.
6. Keep the guide compact while preserving all necessary task phases.
7. Empty lists are allowed when a field has no applicable content.

Return exactly:
{
  "Primary": [
    "..."
  ],
  "Checks": [
    "..."
  ],
  "Avoid": [
    "..."
  ],
  "Fallback": [
    "..."
  ]
}
\end{lstlisting}
\end{tcolorbox}
\caption{Prompt template for the one-time SkillAligner adaptation call.}
\label{fig:skill_adaptation_prompt}
\end{figure}

For ALFWorld, the benchmark-specific execution specification describes the household action interface and asks SkillAligner to preserve the necessary phases of object search, receptacle access, required state transformation, final placement, and progress validation when supported by the retrieved skills. For WebShop, it describes the visible \texttt{search} and \texttt{click} interface and emphasizes search, product filtering, evidence inspection, option binding, hard-attribute verification, price verification, and purchase readiness. For the search-augmented QA tasks, retrieved skills are treated only as reasoning guidance: SkillAligner may preserve decomposition, answer-type, and verification strategies, but must not treat skill content as evidence for the answer.

\section{Regression Source Evaluation}
\label{app:regression_source_evaluation}

We use a GPT-5.5-based LLM judge to identify the primary source of each regression included in the diagnostic analysis. The diagnostic set underlying Table~\ref{tab:regression_sources} consists of all paired instances in which No Skill succeeds but Top-$k$ Raw Skill fails. For every such instance, the judge receives the task query, execution specification, retrieved skills, and paired execution traces. It assigns the regression to one of four mutually exclusive categories: inter-skill misalignment, skill--environment misalignment, skill--task misalignment, or other. The judge is used only for post-hoc error analysis and does not affect skill retrieval, adaptation, execution, or benchmark scoring. Figure~\ref{fig:regression_source_prompt} presents the complete classification prompt.

\section{Additional Results}

\begin{table}[t]
\centering
\small
\renewcommand{\arraystretch}{1.08}
\begin{tabular*}{\textwidth}{@{\extracolsep{\fill}}lcccccc@{}}
\toprule
\multirow{2}{*}{\textbf{\# Skills}}
& \multicolumn{3}{c}{\textbf{Qwen3-8B}}
& \multicolumn{3}{c}{\textbf{Qwen3-32B}} \\
\cmidrule(lr){2-4} \cmidrule(lr){5-7}
& \textbf{ALFWorld} & \textbf{WebShop} & \textbf{SearchQA}
& \textbf{ALFWorld} & \textbf{WebShop} & \textbf{SearchQA} \\
\midrule
200
& 79.85
& 37.40
& 37.21
& 87.31
& 42.40
& \textbf{44.53}
\\

500
& 79.10
& 38.60
& 36.98
& \textbf{89.55}
& 41.20
& 44.41
\\

1,000
& 80.60
& 37.60
& \textbf{37.37}
& 88.06
& 42.80
& 44.36
\\

2,000
& \textbf{82.84}
& \textbf{39.40}
& 37.25
& 87.31
& \textbf{43.20}
& 44.47
\\
\bottomrule
\end{tabular*}
\caption{\textbf{Effect of skill-library size.}
We evaluate SkillAligner with libraries containing 200, 500, 1,000, and 2,000 skills using Qwen3-8B and Qwen3-32B across ALFWorld, WebShop, and the search-augmented QA suite.
Bold values indicate the best result in each column.}
\label{tab:skill_library_scale}
\end{table}

\subsection{Sensitivity to Skill-Library Scale}
\label{app:skill_library_scale}

We examine whether \method{} is sensitive to the scale of the underlying skill library, since practical skill libraries may vary substantially in size as procedural knowledge accumulates. For this analysis, we construct 200-, 500-, 1,000-, and 2,000-skill variants using the same sources, normalization format, and filtering protocol described in Appendix~\ref{app:skill_library}; the 1,000-skill variant is the primary setting used elsewhere. Table~\ref{tab:skill_library_scale} reports the results. Performance varies non-monotonically with library size, and the best scale depends on both the benchmark and backbone. The 2,000-skill library performs best on Qwen3-8B ALFWorld and WebShop and Qwen3-32B WebShop, whereas smaller libraries are best in the remaining settings. No scale uniformly dominates the others. Overall, \method{} remains effective across the evaluated library sizes and does not rely on a specific scale.

\subsection{Detailed Module Ablation}
\label{app:detailed_module_ablation}

Table~\ref{tab:detailed_module_ablation} expands the main ablation in Table~\ref{tab:main_results} to all \(2^3\) combinations of task grounding, execution-environment alignment, and skill composition. The all-empty row corresponds to Top-\(k\) Raw Skill, while the single-stage and complete rows reproduce the corresponding results in Table~\ref{tab:main_results}. The pairwise rows expose how the stages complement one another rather than evaluating each stage only in isolation.

\begin{table}[t]
\centering
\scriptsize
\setlength{\tabcolsep}{3.1pt}
\renewcommand{\arraystretch}{1.02}
\begin{tabular*}{\columnwidth}{@{\extracolsep{\fill}}ccc|ccc@{}}
\toprule
\multicolumn{3}{c|}{Enabled Modules}
& \multicolumn{3}{c}{Qwen3-8B} \\
\cmidrule(lr){1-3}\cmidrule(lr){4-6}
Ground & Align & Compose & ALFWorld & WebShop & SearchQA \\
\midrule
& & & 61.94 & 29.80 & 27.86 \\
\midrule
\checkmark & & & 66.42 & 34.80 & 32.74 \\
& \checkmark & & 67.91 & 32.40 & 31.46 \\
& & \checkmark & 70.90 & 33.60 & 29.41 \\
\midrule
\checkmark & \checkmark & & 72.39 & 37.00 & 35.86 \\
\checkmark & & \checkmark & 76.12 & 37.20 & 34.47 \\
& \checkmark & \checkmark & 77.61 & 36.60 & 33.58 \\
\midrule
\checkmark & \checkmark & \checkmark & \textbf{80.60} & \textbf{37.60} & \textbf{37.37} \\
\midrule
\multicolumn{3}{c|}{Enabled Modules}
& \multicolumn{3}{c}{Qwen3-32B} \\
\cmidrule(lr){1-3}\cmidrule(lr){4-6}
Ground & Align & Compose & ALFWorld & WebShop & SearchQA \\
\midrule
& & & 73.13 & 36.20 & 33.05 \\
\midrule
\checkmark & & & 76.87 & 39.40 & 39.61 \\
& \checkmark & & 78.36 & 37.80 & 37.92 \\
& & \checkmark & 80.60 & 38.60 & 34.88 \\
\midrule
\checkmark & \checkmark & & 82.09 & 41.00 & 43.72 \\
\checkmark & & \checkmark & 84.33 & 41.80 & 42.06 \\
& \checkmark & \checkmark & 85.82 & 40.60 & 40.91 \\
\midrule
\checkmark & \checkmark & \checkmark & \textbf{88.06} & \textbf{42.80} & \textbf{44.36} \\
\bottomrule
\end{tabular*}
\caption{\textbf{Detailed module ablation of SkillAligner.}
A checkmark indicates an enabled module; an empty cell indicates removal. ALFWorld and WebShop report success rate; the search-augmented QA tasks report exact match.}
\label{tab:detailed_module_ablation}
\end{table}

\mypar{Progressive gains from combining stages.}
Each individual stage improves over raw skill injection in all six benchmark--backbone settings, and the complete system outperforms every single-stage variant. Among all \(2^3\) configurations, the complete system is best in all six settings. Relative to the strongest pair for each benchmark, it gains \(2.99/0.40/1.51\) points on ALFWorld, WebShop, and search-augmented QA, respectively, with Qwen3-8B and \(2.24/1.00/0.64\) points with Qwen3-32B. This consistent advantage shows that the three stages provide complementary benefits and are most effective when applied jointly.

\mypar{Different benchmarks favor different module pairs.}
The strongest pair is the same across both backbones but differs by benchmark. Environment alignment plus skill composition performs best on ALFWorld, where execution depends on state-aware action procedures. Task grounding plus skill composition is strongest on WebShop, where the agent must preserve product constraints while organizing search and variant-selection skills. Task grounding plus environment alignment is strongest on search-augmented QA, where query intent and interface-compatible evidence collection are central. This pattern explains why the single-stage winner also varies in Table~\ref{tab:main_results} and why combining all three stages is more reliable than choosing one fixed adaptation strategy. Consistent with Table~\ref{tab:regression_reduction}, the complete system improves transfer and reduces regression relative to raw skill injection across all Qwen settings.

\begin{figure}[t]
\centering
\begin{tcolorbox}[
    width=\textwidth,
    left=1.2mm,
    right=1.2mm,
    top=1.2mm,
    bottom=1.2mm,
    colback=black!3!white,
    colframe=black,
    title={Regression-source classification prompt template.}
]
\begin{lstlisting}[
    basicstyle=\scriptsize\ttfamily,
    columns=fullflexible,
    breaklines=true,
    breakautoindent=false,
    breakindent=2ex,
    keepspaces=true,
    showstringspaces=false
]
[SYSTEM]
You are evaluating the primary source of a skill-induced regression in a skill-augmented agent.

A skill-induced regression is an instance in which the agent succeeds without skills but fails after directly injecting retrieved skills.

Identify the primary causal mismatch introduced by the retrieved skills. Base the classification only on the provided task, execution specification, retrieved skills, and paired execution traces.

Classify the regression into exactly one category:

Inter-skill:
The failure is primarily caused by an interaction among multiple retrieved skills, such as conflicting procedures, incompatible assumptions, duplicated or competing subgoals, or an inconsistent execution order.

Environment:
A retrieved skill relies on an action, tool, resource, object, permission, state, or interface that is unavailable or unsupported by the execution specification.

Task:
A retrieved skill is semantically related to the task but conflicts with the specific goal, input, hard constraint, expected output, or success condition.

Other:
The evidence is insufficient to attribute the failure to one of the three mismatch types, or the failure is better explained by an unrelated reasoning or execution error.

Evaluation rules:
1. Compare the successful no-skill trace with the failed raw-skill trace and identify how skill injection changed the execution.
2. Do not classify a generic reasoning or action error as a skill mismatch unless it can be linked to retrieved skill content.
3. When several mismatches are present, select the one that most directly caused the final failure.
4. Prefer an upstream causal mismatch when it clearly triggered later errors.
5. Classify the violated relation rather than only the surface error.
6. Use Other when no single category is sufficiently supported.
7. Keep the reason and evidence concise and concrete.
8. Return strict JSON only, without markdown or additional commentary.

[USER]
Task query:
<TASK_QUERY>

Execution specification:
<AVAILABLE_TOOLS_ACTIONS_RESOURCES_AND_CONSTRAINTS>

Retrieved skills:
<RETRIEVED_SKILL_TEXTS>

Successful execution trace without skills:
<NO_SKILL_TRACE>

Failed execution trace with raw skill injection:
<RAW_SKILL_TRACE>

Return exactly:
{
  "primary_source": "Inter-skill",
  "reason": "One concise sentence describing the primary causal mismatch.",
  "evidence": "The relevant skill instruction or paired-trace difference.",
  "confidence": "High"
}

The value of "primary_source" must be exactly one of:
"Inter-skill", "Environment", "Task", or "Other".

The value of "confidence" must be exactly one of:
"High", "Medium", or "Low".
\end{lstlisting}
\end{tcolorbox}
\caption{Prompt used by the LLM judge to classify the primary source of each skill-induced regression.}
\label{fig:regression_source_prompt}
\end{figure}

\section{Detailed Case Studies}
\label{app:detailed_case_studies}

This section expands the main-paper example and adds two complementary WebShop cases. All cases use Qwen3-32B. For each task, No Skill and Top-$k$ Raw Skill use the exact same shopping instruction and evaluation protocol; Raw Skill receives the retrieved top-five skills directly, while SkillAligner adapts the same five skills once before execution. The three cases isolate different failure modes: \textbf{incomplete option binding}, \textbf{negative transfer from an over-general skill}, and \textbf{repetition without a recovery strategy}. WebShop returns a score in $[0,1]$, but its success condition is exact: \textbf{score 1 is success; every score below 1 is failure}. We show sub-unit scores only as diagnostic evidence of partially satisfied constraints.

\begin{table}[h]
\centering
\small
\setlength{\tabcolsep}{3.5pt}
\begin{tabular}{@{}lccc@{}}
\toprule
\textbf{Case} & \textbf{No Skill} & \textbf{Raw Skill} & \textbf{SkillAligner} \\
\midrule
Banana chips & Fail (.25) & Fail (.75) & \textbf{Success (1)} \\
Sound column & \textbf{Success (1)} & \textbf{Fail (.50)} & \textbf{Success (1)} \\
Sea-salt shaker & Fail (0) & Fail (0) & \textbf{Success (1)} \\
\bottomrule
\end{tabular}
\caption{\textbf{Outcomes of the detailed WebShop cases.} Scores below 1 are failures under the success criterion.}
\label{tab:detailed_case_outcomes}
\end{table}

\subsection{Main Case: Configuring a Hidden Product Variant}

\mypar{What makes the task hard.}
The user requests freeze-dried banana chips with \emph{apple cinnamon} flavor, a \emph{16-pack}, and a price below \$50. The search page, however, shows only a default configuration. The correct parent product B092JLLYK6 is titled as generic banana crisps in a 6-pack; only after opening its detail page does the agent see two separate controls for flavor and pack size. This is analogous to a shoe listing that displays one default color and size even though other variants are selectable. \textbf{The search title is therefore not the final SKU: the agent must open the parent product and explicitly satisfy both option groups.}

\mypar{No Skill: constraints are not tracked.}
The No-Skill agent enters a different fruit-crisp family, selects \emph{goofy strawberry banana} and \emph{pack of 24}, and buys it:
\begin{center}
\small
\texttt{wrong fruit crisps} $\rightarrow$ \texttt{strawberry banana}\\[-1mm]
$\rightarrow$ \texttt{pack of 24} $\rightarrow$ \texttt{Buy Now}.
\end{center}
The score is 0.25, hence failure. The agent clearly knows how to search and click an option. Its problem is simpler and more specific: \textbf{it does not treat flavor and pack size as two requirements that must both remain satisfied before purchase}.

\mypar{Raw Skill: three mismatches accumulate.}
The retrieved skills cover query parsing, packaging-aware search, search recovery, candidate selection, and variant choice, but direct injection leaves three failures unresolved. \textbf{Inter-skill mismatch:} after a candidate is rejected, the Search Executor returns to the results while the Product Selector chooses the same literal match again, producing a six-round loop instead of switching candidates. \textbf{Environment mismatch:} the Variant Chooser treats one option click as a complete configuration; it selects the 16-pack but leaves the independent apple-cinnamon flavor unset. \textbf{Task mismatch:} the Product Selector prioritizes query relevance without enforcing the price-below-\$50 requirement, so the final purchase is over budget. \textbf{The skills are individually useful, but direct injection supplies neither a coherent recovery handoff nor a purchase gate covering every hard constraint.} The resulting score is 0.75 and therefore counts as a failure.

\mypar{SkillAligner: track four hard slots until purchase.}
\textbf{Ground} rewrites the request as four independent slots: banana chips, apple cinnamon, 16-pack, and price below \$50. \textbf{Align} records that search results display default variants and that flavor and pack become verifiable only through separate detail-page clicks. \textbf{Compose} merges the overlapping search instructions, forbids revisiting an already rejected literal match, orders the two option clicks, and allows purchase only when all four slots pass. The resulting route is
\begin{center}
\small
\texttt{search} $\rightarrow$ \texttt{configurable parent} $\rightarrow$ \texttt{apple cinnamon}\\[-1mm]
$\rightarrow$ \texttt{pack of 16} $\rightarrow$ \texttt{price check} $\rightarrow$ \texttt{Buy Now},
\end{center}
which receives a score of 1.

\mypar{What this case isolates.}
Raw Skill already proves that the model can form a useful query, reject a bad product, inspect candidate details, and click a legal variant. \textbf{The missing capability is not search or tool use; it is turning several locally useful skills into one complete, state-aware shopping procedure.} SkillAligner supplies the missing handoff: reject once, change candidate, bind flavor, bind pack, check price, then buy.

\subsection{Supplementary Case I: Preventing Negative Transfer}

\mypar{What makes the task diagnostic.}
The user requests a high-power \emph{sound column} subwoofer with Bluetooth and 3D surround sound for less than \$660. Here, ``sound column'' names the requested product type; it is not an optional descriptive word. The retrieved skills cover search formulation, result filtering, product inspection, and purchase. This case is especially clean because \textbf{the backbone succeeds without skills but fails after Raw Skill is injected}.

\mypar{No Skill: product type preserved.}
Without skills, the agent keeps the full category phrase in its query:
\begin{center}
\small\ttfamily
high power sound column subwoofer\\[-1mm]
bluetooth 3d surround
\end{center}
The result list contains B09R1DKTS6, whose visible title states sound column, high power, Bluetooth, and 3D surround, at \$625.85. The agent opens it and purchases it in three valid actions, receiving a score of 1.

\mypar{Raw Skill: task-defining term dropped.}
The search-formulation skill recommends using only a few distinguishing attributes. Applied without task grounding, this generic rule removes \emph{sound column}:
\begin{center}
\small\ttfamily
high power subwoofer bluetooth\\[-1mm]
3d surround sound
\end{center}
This shortening changes which products appear. The agent selects B093V43SQM, an ordinary TV soundbar/subwoofer, and immediately purchases it. Every action is legal, but the product type is wrong, so the score is 0.5 and the run fails. \textbf{The skill does not fail because query shortening is always harmful; it fails because it treats the task-defining category as removable.}

\mypar{SkillAligner: protect the category before shortening the query.}
\textbf{Ground} marks \emph{sound column} as non-droppable and keeps high power, Bluetooth, 3D surround, and budget as separate checks. \textbf{Align} removes an unavailable selector-script call and retains the comparison rule in a form executable through visible search and click actions. \textbf{Compose} gives query formulation one owner and requires product-type verification before purchase. The aligned query therefore preserves \emph{sound column}, recovers B09R1DKTS6, and receives a score of 1.

\mypar{What this case isolates.}
\textbf{SkillAligner prevents negative transfer rather than merely helping an otherwise incapable agent.} The paired No-Skill success shows that search, product selection, and purchase are already within the backbone's ability; the only damaging change is the Raw-Skill rewrite that drops the hard category.

\subsection{Supplementary Case II: Escaping a Literal-Match Loop}

\mypar{What makes the task hard.}
The user requests a one-pound, pack-of-one organic sea-salt shaker in the \emph{triple blend flakes} flavor for less than \$30. The first result page contains two competing candidates. B01GGWDB8S looks best by title because it explicitly says ``Triple Blend Flakes,'' but it is a one-ounce two-pack priced at \$100. B0007SMLUM has the less literal title ``Sea Seasoning Shakers---Organic'' and costs \$6.89; opening it reveals the requested flavor and one-pound pack-of-one options. \textbf{The agent must reject the best lexical match and inspect the less literal parent product.}

\mypar{No Skill: rejected candidate repeated.}
The agent opens B01GGWDB8S, sees the incompatible size, pack, and \$100 price, returns to the result page, and opens the same product again. This cycle consumes the full 30-step interaction budget:
\begin{center}
\small
\texttt{search} $\rightarrow$ \texttt{\$100 literal match} $\rightarrow$ \texttt{reject}\\[-1mm]
$\rightarrow$ \texttt{same search} $\rightarrow$ \texttt{same product}.
\end{center}
The score is 0. \textbf{The failure is not detecting that the product is invalid; it is failing to change strategy after that detection.}

\mypar{Raw Skill: no recovery handoff.}
The top-five bundle contains a query parser, result filter, search executor, variant chooser, and purchase gate. In isolation, these skills cover the full solution: extract each constraint, reject invalid products, open a parent product, select flavor and size, then buy. Nevertheless, Raw Skill issues the same query eight times and repeatedly reopens the same \$100 listing. The filter rejects the product, but the search executor is not told to exclude it or choose a different candidate; therefore the variant chooser never receives a usable parent page. \textbf{Retrieving the complete set of local procedures is insufficient when no procedure owns the handoff after rejection.}

\mypar{SkillAligner: reject once, inspect another candidate.}
\textbf{Ground} stores product form, flavor, one-pound size, pack count, and budget as independent checks, so a title match cannot hide three visible violations. \textbf{Align} activates the variant chooser only on a detail page with option controls and drops an unavailable parser-script call. \textbf{Compose} turns rejection into an explicit next step: after rejecting B01GGWDB8S once, the fallback must inspect another candidate, B0007SMLUM, rather than repeat the same pair. It then orders
\begin{center}
\small
\texttt{B0007SMLUM} $\rightarrow$ \texttt{triple blend flakes}\\[-1mm]
$\rightarrow$ \texttt{1 pound (pack of 1)} $\rightarrow$ \texttt{Buy Now}.
\end{center}
The public product, option, and price checks pass, and the run receives a score of 1. Organic or certification claims remain public-page checks; the causal comparison does not rely on hidden evidence, because the rejected literal listing already violates size, pack, and budget.

\mypar{What this case isolates.}
\textbf{The bottleneck is executable organization, not missing knowledge.} Raw Skill already contains every major step, but SkillAligner is needed to connect rejection to a new candidate and to activate variant selection only after the correct parent page is open.

\subsection{Cross-Case Interpretation}

The cases expose three gaps: \textbf{(i) incomplete state}---the banana-chip agent buys after one option; \textbf{(ii) harmful rewriting}---query shortening drops the sound-column product type; and \textbf{(iii) stalled recovery}---the sea-salt agent revisits a rejected SKU. Accordingly, \textbf{Ground protects hard constraints}, \textbf{Align binds procedures to visible state}, and \textbf{Compose assigns the next owner, purchase gate, and fallback}---the links missing from Raw Skill.

\end{document}